\documentclass[twocolumn,superscriptaddress]{revtex4-1}
\usepackage{amssymb}
\usepackage{mathtools}
\usepackage{color}
\usepackage{algorithm}
\usepackage{algorithmic}
\usepackage[colorlinks=true,urlcolor=blue,citecolor=blue]{hyperref}

\newcommand{\tr}{\mathrm{tr}}

\newcommand{\lspan}{\mathrm{span}}

\newtheorem{prop}{Proposition}

\begin{document}
\title{Quantum Enhanced Inference in Markov Logic Networks}

\author {Peter Wittek}
\affiliation{ICFO-The Institute of Photonic Sciences, 08860 Castelldefels (Barcelona), Spain}
\affiliation {University of Bor{\aa}s, 50190 Bor{\aa}s, Sweden}
\author {Christian Gogolin}
\affiliation{ICFO-The Institute of Photonic Sciences, 08860 Castelldefels (Barcelona), Spain}

\begin{abstract}
Markov logic networks (MLNs) reconcile two opposing schools in machine learning and artificial intelligence: causal networks, which account for uncertainty extremely well, and first-order logic, which allows for formal deduction. An MLN is essentially a first-order logic template to generate Markov networks. Inference in MLNs is probabilistic and it is often performed by approximate methods such as Markov chain Monte Carlo (MCMC) Gibbs sampling. An MLN has many regular, symmetric structures that can be exploited at both first-order level and in the generated Markov network. We analyze the graph structures that are produced by various lifting methods and investigate the extent to which quantum protocols can be used to speed up Gibbs sampling with state preparation and measurement schemes.
We review different such approaches, discuss their advantages, theoretical limitations, and their appeal to implementations. We find that a straightforward application of a recent result yields exponential speedup compared to classical heuristics in approximate probabilistic inference, thereby demonstrating another example where advanced quantum resources can potentially prove useful in machine learning.
\end{abstract}
\maketitle
\section{Introduction}
Graphical models combine uncertainty and logical structure in an intuitive representation.
Examples include Bayesian networks, Markov networks, conditional random fields, and hidden Markov models, but also Ising models and Kalman filters.
Their main advantage is the compactness of representation, stemming from capturing the sparsity structure of the model and independence conditions among the variables reflected in the correlations.
The graph structure encompasses the qualitative properties of the distribution.
Exact probabilistic inference in a general Bayesian or Markov network is \#P-complete~\cite{koller2007graphical},
which is why one often resorts to  Markov chain Monte Carlo (MCMC) Gibbs sampling to approximate exact probabilistic inference.
However, the task remains computationally intensive even with MCMC.

Graphical models belong to a school of machine learning that emphasizes the importance of probability theory.
First-order logic on the contrary comes from the symbolist tradition of artificial intelligence and it relies on inverse deduction to perform inference.
Markov logic networks reconcile the two schools, and in one limit, they recover first-order logic~\cite{richardson2006markov}.
A Markov logic network is essentially a template for generating Markov networks based on a knowledge base of first-order logic.
MCMC Gibbs sampling can be used in the same way as in ordinary Markov networks to perform approximate probabilistic inference, but it suffers from the enormous number of nodes that are generated by the template.

There has been a recent surge of interest in using quantum resources to improve the computational complexity of various tasks in machine learning~\cite{schuld2014introduction,wittek2014qml,adcock2015advances}.
This approach has been successful in training Boltzmann machines, which are simple generative neural networks of a bipartite structure---a set of hidden and a set of visible nodes---where the connectivity is full between the two layers.
Edges carry weights and these are adjusted during training.
We can view Boltzmann machines as Markov networks with a special topology, in which the largest clique has size two.
One method employed for training Boltzmann machines~\cite{adachi2015application,benedetti2015estimation,perdomo-ortiz2015determination} is quantum annealing.
It is a global optimization method that relies on actual physical phenomena and it can be used to generate a Gibbs distribution.
For all current quantum annealing approaches to Gibbs sampling, restrictions on the topology of the physical hardware remain the main obstacle, which is why the limited clique size of the Boltzmann machines is attractive.
An alternative approach of training Boltzmann machines is by using Gibbs state preparation and sampling protocols, which can also exploit the structure of the graph and achieve polynomial improvements in computational complexity relative to its classical analogue~\cite{wiebe2014quantum}.

Here, we go beyond the training of Boltzmann machines and consider more general Markov logic networks, keeping the expressiveness of first-order logic and concentrate on inference, rather than training.
We analyze the usefulness of quantum Gibbs sampling methods to outperform MCMC methods.
The runtime of quantum Gibbs sampling algorithms is sensitive to both to the connectivity structure and the overall number of subsystems.
Methods of lifted inference can be used to address these issues.

\section{Probabilistic Inference and Lifting}
\begin{table*}
\begin{center}
\begin{tabular}{l|l}
  \textbf{First-order formula} & \textbf{Graph characteristic}\\
  \hline
  Number of atoms in formulas & Clique size\\
  Domain size and number of atoms in formula & Total number of nodes \\
  Maximum shared variables & Largest degree
\end{tabular}
\end{center}
\caption{Brief summary of how the structure of the first-order formulas in the knowledge base underlying a Markov logic network influences the generated Markov network.
Shared variables are variables that appear in more than one formula.}
\label{foltograph}
\end{table*}
Markov networks are undirected graphical models that offer a simple perspective on the independence structure of a joint probability distribution of random variables, and the task of probabilistic inference based on this structure~\cite{koller2007graphical}.
Nodes of the network are random variables and edges between nodes imply influence or direct correlation, that is, lack of conditional independence.
Instead of conditional probabilities on parent nodes, as in Bayesian networks, Markov networks operate with unnormalized \emph{factors} $f_j$, that is, functions that map from subsets of the random variables to nonnegative reals.
The factors are defined over the cliques of the graph.
To obtain a valid joint probability distribution over the random variables from the factors, a partition function normalizes the unnormalized measure, so that the probability distribution takes the form $P(\mathbf{X}=\mathbf{x}) = \frac{1}{Z}\prod_{j} f_j(\mathbf{x}_j)$, where $\mathbf{x}_j$ are subsets of $\mathbf{x}$ corresponding to the cliques and $Z$ is the partition function.
If $P$ is a positive distribution over the random variables $\mathbf{X} \coloneqq (X_1,\ldots,X_n)$, we can associate a Gibbs distribution to the Markov network as $P(\mathbf{X}=\mathbf{x}) = \frac{1}{Z}\exp(\sum_{j} w_j\, g_j(\mathbf{x}))$, where the \emph{features} $g_j$ are functions of a subset of the state, and $w_j$ are real weights.

In first-order logic, \emph{constants} are objects over some \emph{domain} (e.g., \texttt{Alice}, \texttt{Bob},\ldots in the domain of people), and \emph{variables} range over the set of constants in the domain.
A \emph{predicate} is a symbol that represents an attribute of an object (e.g, \texttt{Smokes}), or a relation among objects (e.g., \texttt{Friends}).
An \emph{atom} is a predicate applied to a tuple of variables or constants.
A \emph{ground atom} only has constants as arguments.
These definitions apply to a function free language with finite size domains---technically, this is a strict subset of first-order logic.
A \emph{formula} is constructed of atoms, logical connectives, and quantifiers over variables.
A \emph{knowledge base} is a set of formulas connected by conjunction.
A \emph{world} is an assignment of a truth values to each possible grounding of all atoms in a knowledge base.
An essential task in a first-order knowledge base is to check whether a formula is \emph{satisfiable}, that is, there exists at least one world in which it is true.

To relax the rigid true-or-false nature of first-order logic, Markov logic networks (MLNs) introduce a real weight $w_j$ for each formula $f_j$ in a knowledge base~\cite{richardson2006markov}.
A Markov logic network $\mathcal{M}$ is a set of pairs $(f_j, w_j)$, representing a probability distribution over worlds as
\begin{equation} \label{eq:probability_MLN}
  P_\mathcal{M}(\omega) \coloneqq \frac{1}{Z(\mathcal{M})}\exp\left(\sum_j w_j \,N(f_j, \omega)\right) ,
\end{equation}
where $N(f_j, \omega)$ is the number of groundings of $f_j$ that are $\mathtt{True}$ in the world $\omega$.
An MLN can be thought of as a graph over the set of all possible groundings of the atoms appearing in the knowledge base.
The size of this graph is $n \in \mathcal{O}(D^c)$, where $D$ is the maximum domain size, and $c$ is the highest number of atoms in any of the formulas in the knowledge base~\cite{kersting2009counting}.
Groundings are viewed as connected if they can jointly appear in a grounding of some formula of the knowledge base.
The ground network thus contains cliques, i.e., fully connected sub-graphs, consisting of grounded atoms that jointly appear in the grounding of some formula.
The maximum clique size $k$ is given by the maximum number of atoms per formula.
Table~\ref{foltograph} summarizes how the structure of the first-order knowledge base influences the characteristics of the generated Markov network.

MLNs belong to the class of methods known as statistical relational learning, which combine relational structures and uncertainty~\cite{getoor2007introduction}.
An MLN essentially uses a first-order logic knowledge base as a template to generate a Markov network by grounding out all formulas.
An MLN can always be converted to a \emph{normal} MLN, which has the following two properties: (i) there are no constants in any formula; (ii) given two distinct atoms with the same predicate symbol with two variables $x$ and $y$ in the same argument, then the domain of the two variables is identical.
In the rest of this work we assume all MLNs to be given in this normal form.
We further assume that skolemnization is applied to convert existential quantifiers to universal quantifier, which can be done in polynomial time in the size of a formula with no unquantified variables~\cite{vandenbroeck2014skolemnization}.

A main task in graphical models and in MLNs is probabilistic inference.
One aspect of it is computing the partition function.
The other aspect deals with the problem of assigning probabilities to or finding (the most) likely assignment of variables given evidence, that is, given a fixed assignment for a subset of its variables.
This is a hard problem in general: the worst-case complexity of exact probabilistic inference of a graphical model is $\#\mathcal{P}$-complete and that of approximate inference is $\mathcal{NP}$-hard~\cite{koller2007graphical}.

For some common graphical models with a special topology, efficient exact probabilistic inference methods are known.
Examples include belief propagation~\cite{pearl1982reverend} and the junction tree algorithm~\cite{lauritzen1990local}.
In other cases, MCMC Gibbs sampling is often used for approximate inference to escape the worst-case complexity of exact inference.
MCMC is hereby used to approximately sample from the distribution $P_\mathcal{M}(\omega)$ given in \eqref{eq:probability_MLN} or from a suitable conditional probability distribution $P_\mathcal{M}(\omega|\mathcal{E})$ conditioned on the evidence $\mathcal{E}$.

Graphical models often have symmetries that reduce the overall complexity of both exact and approximate inference.
For instance, counting belief propagation exploits symmetries for exact inference~\cite{kersting2009counting}, and orbital Markov chains do the same for approximate inference~\cite{niepert2012markov}.
Some of these methods have special extensions for MLNs, for instance, one can detect a subset of components in the ground network that would behave identically during belief propagation~\cite{singla2008lifted}.
It is worth exploiting the symmetries that emerge from first-order logic and they are best exploited before grounding out, that is, symmetries should be addressed at the propositional level.

Approximate and exact probabilistic inference for first-order probabilistic languages predates MLNs~\cite{pasula2001approximate,poole2003first,desalvobraz2005lifted}.
The core idea is a form of coarse graining by grouping similar variables together.
This idea was exploited in lifted first-order probabilistic inference for MLNs~\cite{kersting2012lifted}.
For hierarchically typed MLNs, one can move from coarse-graining over the highest level in a type hierarchy to more refined types~\cite{kiddon2011coarse}.

Exploiting symmetries in the presence of evidence must be done with great care.
Given evidence, the symmetries can become skewed, as random variables do not appear symmetrically in the formulas of the knowledge base~\cite{ahmadi2013exploiting}.
In this case, importance sampling helps~\cite{gogate2012action,venugopal2014scalingup}, which clusters similar network components together given the evidence~\cite{venugopal2014evidence}, and approximates the correct probabilities by an easier probability distribution and an estimated importance or weight of the error.

For most practical applications, either belief propagation or MCMC, augmented with some of the described techniques as appropriate for the problem at hand, is the method of choice for approximate probabilistic inference with MLNs.
While often yielding useful results with an effort far smaller than the worst case complexity, they remain very expensive computationally and so more efficient alternatives are desirable.

\section{Quantum Gibbs Sampling}
\begin{figure*}[htb!]
	\centering
	\includegraphics[width=\textwidth]{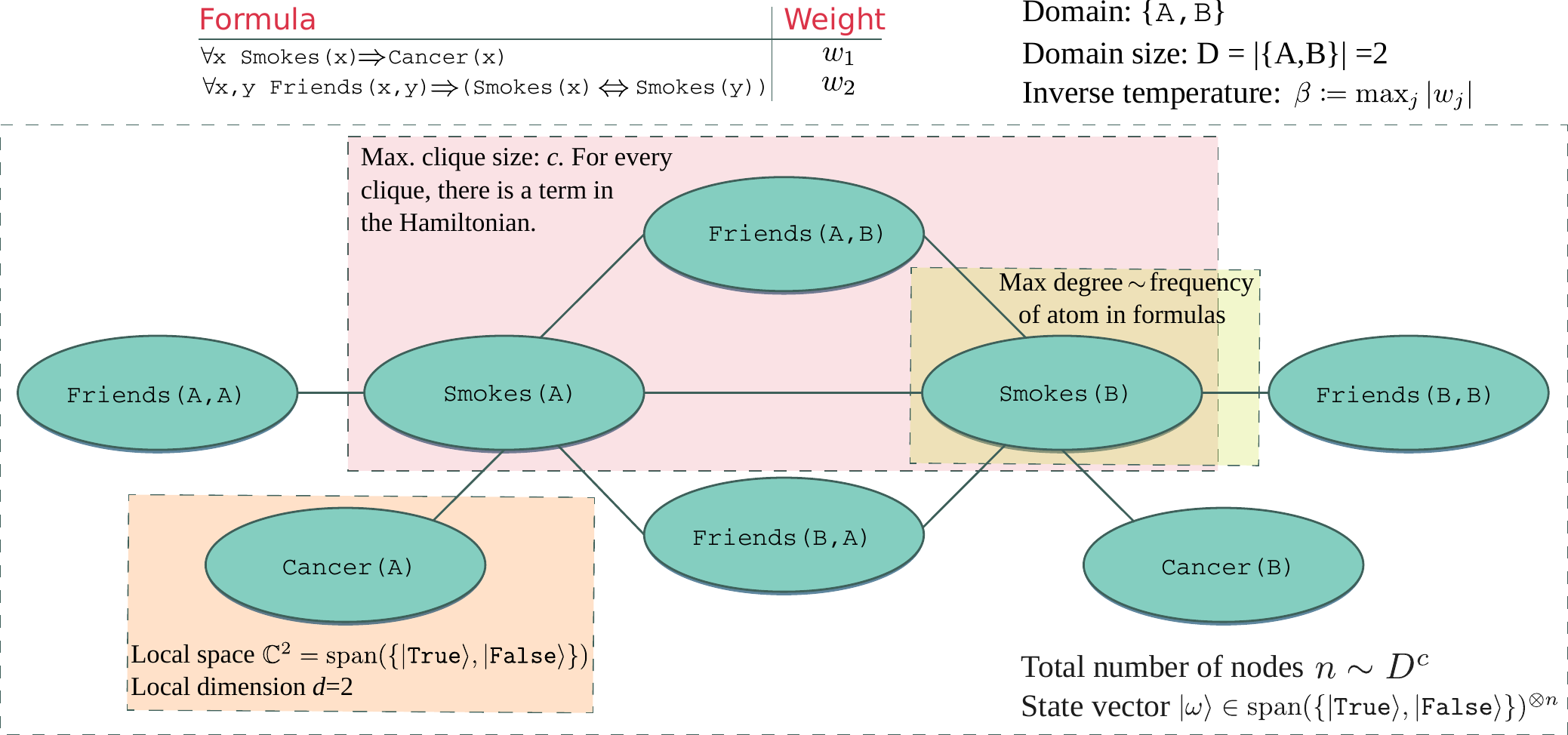}
	\caption{An example of a first-order knowledge base, a matching MLN, and the corresponding concepts of a thermal state and a local Hamiltonian. The knowledge base has only two formulas, and the variables range over a finite domain of two elements, $\{A, B\}$. Grounding out all formulas in all possible way, we obtain the MLN of maximal size (i.e., lifted inference is not used). The maximum of absolute value of the weights $w_1$ and $w_2$ defines the inverse temperature $\beta$ in the thermal state. Since all ground atoms are binary valued, the local space is $\mathbb{C}^2=\lspan(\{|\mathtt{True}\rangle, |\mathtt{False}\rangle\})$, and thus the thermal state $|\omega\rangle$ is in $\lspan(\{|\mathtt{True}\rangle, |\mathtt{False}\rangle\})^{\otimes n}$, where $n$ is the total number of nodes.}
	\label{qmln_concepts}
\end{figure*}

The distribution \eqref{eq:probability_MLN} we would like to sample from can be thought of as the Gibbs distribution of a suitably constructed physical system.
According to the rules of statistical mechanics, the probability to find a system in a certain state of configuration when it is in thermal equilibrium follows a Gibbs distribution.
The distribution can thus be sampled by preparing a suitable physical system in a thermal equilibrium Gibbs state and then measuring its configuration.
This is generally rather easy to do at high temperatures, but cooling to low temperatures typically becomes increasingly difficult.
Thereby methods of quantum information processing can offer advantages over classical strategies.
They open up fundamentally new ways to of preparing systems approximately in Gibbs states in a well-controlled way.

Going from the abstract definition of the probability distribution in \eqref{eq:probability_MLN} to a physical model that can be done in the following way:
We can think of $-\sum_j w_j\, N(f_j, \omega) / \max_j |w_j|$ as the ``energy'' of a system of $n$ spin $1/2$ ``particles'' in the quantum state $|\omega\rangle$.
The states $|\omega\rangle$ are then product state vectors in the Hilbert space $\lspan(\{|\mathtt{True}\rangle, |\mathtt{False}\rangle\})^{\otimes n}$ with $\lspan$ the complex linear span.
We can think of $\beta \coloneqq \max_j |w_j|$ as the inverse $\beta = 1/(k_B\,T)$ of the ``temperature'' $T$ of the system time the Boltzmann constant $k_B$ (other decompositions of the features are also possible).
We can try to find a Hamiltonian $H$ such that we can rewrite the probability distribution from \eqref{eq:probability_MLN} as follows
\begin{equation}
P_\mathcal{M}(\omega) = \langle \omega| \exp(-\beta\,H) / Z(\mathcal{M}) | \omega \rangle .
\end{equation}
Thereby $\langle \omega|$ is the Hermitian conjugate of the state vector and $Z(\mathcal{M}) \coloneqq \tr(\exp(-\beta\,H))$ is the partition function, where $\exp$ is the matrix exponential and $\tr$ the matrix trace.

In the concrete case of an MLN the number of particles $n \in \mathcal{O}(D^c)$ is equal to the number of all possible groundings of the atoms in the knowledge base underlying the MLN.
The Hamiltonian $H$ inherits the locality structure of the MLN:
it can be written as a sum $H = \sum_l h_l$ of local terms $h_l$, one for each clique of the MLN.
More precisely, for each $j$ the expression $N(f_j, \omega)$ translates to a sum over local terms each acting on one of the cliques produced by grounding out $f_j$ and acting on this clique like $-w_j/ \max_j |w_j|$ times the projector on the subspace of assignments to the atoms in the clique for which $f_j$ evaluates to $\mathtt{True}$.
The local terms $h_l$ of the Hamiltonian can be constructed from the truth tables of the the $f_j$ and the sum over $l$ in the decomposition of $H$ collects all such terms for the different values of $j$. 
Figure~\ref{qmln_concepts} illustrates the matching concepts in MLNs and this description.

The number $k$ of subsystems on which each such term acts non-trivially is bounded by the maximum number of atoms per formula and its operator norm is bounded by one $\|h_l\| \leq 1$.
Hence \eqref{eq:probability_MLN} is the thermal Gibbs distribution of a system of $n$ spin $1/2$ particles with a so-called $k$-local Hamiltonian $H$.
To prepare the system in a state that is suitable to sample from \eqref{eq:probability_MLN} it is sufficient to reach a high effective temperature if all weights are of moderate magnitude (no assignments are strongly suppressed), but it is necessary to cool to a low temperature if weights have a high magnitude (at least one assignment is strongly suppressed).

\section{Computational Complexity}
Quantum Gibbs sampling methods can be used to obtain samples from the Gibbs distributions of the type of systems described in the previous section.
Typically these methods consist of two phases:
\begin{enumerate}
	\item A preparation phase in which a quantum system is prepared in (a state close to) a state encoding information about the Gibbs state or such a state itself; and
	\item A measurement phase, in which, by performing measurements on this state, samples from the Gibbs distribution are obtained.
\end{enumerate}

The measurement phase is trivial, consisting only of local measurements and has complexity $\mathcal{O}(n)$.
The known quantum methods for Gibbs sampling differ in the kind of resources they require during the preparation, their expected improvement in runtime over classical methods, and the extend to and effort with which their performance for a concrete Gibbs distribution can be predicted.

The state prepared in the preparation phase is usually either close to a thermal Gibbs state~\cite{poulin2009sampling,chiang2010improvement,riera2012thermalization,chowdhury2016quantum} at inverse temperature $\beta$ of a given Hamiltonian $H$, or to a so-called pure thermal state~\cite{wocjan2009estimatingpartitionfunctions,boixo2009quantumsimulatedannealing}, i.e., a pure state whose overlap with any energy eigenstates of $H$ with energy $E$ is proportional to the square root of the Gibbs weight $\exp(-\beta\,E)$.

Recently, an algorithm for the computation based approximate preparation of thermal states of arbitrary $k$-local Hamiltonians has been proposed in~\cite{chowdhury2016quantum}.
For this algorithm a particularly favorable upper bound on the gate complexity---the scaling of the number of elementary operations in the preparation step---is known.
This bound can be expressed in terms of the inverse temperature $\beta$, the local dimension $d$ ($d=2$ for Gibbs states corresponding to MLNs), the number of local terms in $H$, the gate complexity of time evolution under these terms (or the size of their support) and their strength, as well as the value of the partition function $Z = \tr(\exp(-\beta\,H))$, and the final distance to the thermal state $\epsilon$.

\begin{prop}\label{prop:complexity}
Assuming that the maximum size of the support of the local terms of the Hamiltonian $H$ is constant\footnote{More generally it is enough if the gate complexity of time evolution under the local terms of $H$ scales at most linear with $n$.} and that for some constant $\alpha$ the number of terms in $H$ is in $\mathcal{O}(n^\alpha)$, the overall complexity of the Gibbs sampling method from~\cite{chowdhury2016quantum} is in $\mathcal{O}(\sqrt{d^n\beta/Z}\,\mathrm{polylog}(\sqrt{d^n\beta/Z}/\epsilon))$.
\end{prop}

When applying this to the graph structure generated by an MLN, $\alpha$ can be taken to be the maximum number of atoms in any formula, the maximum size of the supports of the local terms of $H$ is equal to the maximum clique size, and the number of terms is the number of cliques in the MLN.
So long as the maximum number of atoms in any formula is constant, the above scaling of complexity is achieved.
It is important to note that the complexity does not directly depend on the maximal degree of the MLN.

This result improves upon the previously known methods in several respects, but in particular it improves the scaling of the runtime with $1/\epsilon$ and $\beta$.
In the natural parameters, the problem size $n$ and the precision $\epsilon$, this method yields an exponential improvement over the runtime of classical simulated annealing, which scales like $1/(\delta\,\epsilon^2)$, where $\delta$ is the gap of the Markov process, which in interesting cases typically is in $\mathcal{O}(1/d^n)$~\cite{wocjan2009estimatingpartitionfunctions}.
However, the exponential dependence on $n$ remains.
The possibility of a logarithmic scaling with $1/\epsilon$ was anticipated in~\cite{richter2007periodic,somma2008combinatorial,tucci2009gibbssampling}.
This scaling is particularly relevant when small probabilities are to be estimated with small relative error.

Following early works~\cite{terhal2000quantumequilibrium}, several previous methods for quantum Gibbs sampling with improved scaling of complexity had been proposed~\cite{wocjan2009estimatingpartitionfunctions,somma2008combinatorial,somma2007combinatorial,richter2007periodic,poulin2009sampling,chiang2010improvement,tucci2009gibbssampling,wocjan2008speedup,boixo2009quantumsimulatedannealing}.
This in particular concerns the dependence of the runtime on the dimension of the Hilbert space $d^n$ or the inverse gap of a Markov chain $1/\delta$, which was reduced from linear to square root by using techniques such as Szegedy's quantum walks, Grover's algorithm, phase estimation, or amplitude amplification.
Algorithms that speed up the convergence of Markov Chains with quantum techniques~\cite{wocjan2009estimatingpartitionfunctions,somma2008combinatorial,somma2007combinatorial,richter2007periodic,tucci2009gibbssampling,wocjan2008speedup,boixo2009quantumsimulatedannealing} often offer more flexibility than such more specific to the problem of preparing thermal states~\cite{poulin2009sampling,chiang2010improvement,chowdhury2016quantum}. In cases in which the gap a Markov chain is much larger than $1/d^n$, they combine their quantum speedup with the advantage inherent in MCMC.
However, the interesting cases are usually those in which $1/\delta \approx d^n$ and then both types of algorithms perform essentially equally well.
A different method, based on the preparation of microcanonical states, was developed in~\cite{riera2012thermalization}, but has an at least exponential scaling in $\beta\,\|H\|$.

If the Hamiltonian $H$ has more structure and/or the effective temperature is high, more efficient special purpose procedures are available~\cite{kastoryano2016quantum,bilgin2010dimensionreduction,wiebe2014quantum}, which however are of limited relevance for inference in MLNs.
In addition to that, there exists a quantum generalization of the Metropolis sampling algorithm~\cite{temme2011metropolis}, that however does not aim at achieving a speedup, but rather works around the sign problem in fermionic systems and makes MCMC techniques available for general local quantum Hamiltonians with non-commuting terms.

\subsection{Can We Hope for Something Better?}
As we have seen, quantum methods reduce the complexity of approximate Gibbs sampling quite drastically.
Still, an exponential scaling with the number of all possible groundings of all atoms $n$ remains, and the complexity diverges in the low temperature limit as $\beta$ goes to infinity.
A valid question is: Can we hope that future advances will remedy this?
After all, the quantum Gibbs sampling methods presented above are able to do Gibbs sampling from Hamiltonians much more general than those that can arise from MLNs, like ones that have non-commuting terms, for example.
Yet, the answer is probably negative.
It is highly unlikely that any general purpose quantum algorithm for inference in MLNs exists that is efficient in cases with high weights (i.e., at low temperatures), as this would imply the existence of an efficient algorithm for solving satisfiability problems more general than 3-SAT, which is known to be NP-complete by the Cook--Levin theorem~\cite{cook1971sat}.
Further, the $\log(1/\epsilon)$ scaling of complexity is known to be optimal for Hamiltonian simulation~\cite{berry2015simmulation} and hence for any Gibbs sampling method based on it.
The situation is different in the high temperature regime, where more efficient Gibbs sampling methods exist~\cite{kastoryano2016quantum}.

\subsection{Computing the Partition Function with First-Order Lifting}\label{sec:partfunc}
The great advantage of working with lifting at the first-order level is that we save potentially exponentially many groundings given the compact representation when we count the models in Eq.~(\ref{eq:probability_MLN}). There are trivial cases: for instance, when there are no shared variables between the atoms, then there is a closed form to calculate the number of satisfied groundings~\cite{sarkhel2014lifted}. Here we follow the outlines of lifted importance sampling~\cite{gogate2011probabilistic,gogate2012action}, but without reference to an importance or proposal distribution: our aim is to reduce the complexity of the generated Markov network and potentially split it into disconnected graphs when computing the partition function. We run quantum Gibbs sampling on the smaller network and post-process the result with some book-keeping values to return the value of the partition function. Algorithm~\ref{alg:liftedmln} summarizes the steps. Since the sampling is not based on a proposal distribution, the actual variance will depend on the error term that estimates the accuracy of the quantum Gibbs sampler. We follow the simplification steps from lifted importance sampling to cater to the critical parts of quantum thermal state preparation, but in principle, the sampling part of the algorithm can also use classical MCMC Gibbs sampling. For this reason, Algorithm~\ref{alg:liftedmln} does not specify what kind of Gibbs sampling protocol we use.

If we have a normal-form network as the input, that is, all domains have size one, we can run the Gibbs sampler and return the value of the partition function.

The first interesting case is if we detect a \emph{decomposer}---this can be done in linear time---that is, a set of logical variables $\mathbf{x}$ such that (i) every atom in $\mathcal{M}$ contains exactly one variable from $\mathbf{x}$, and (ii) for every predicate $\mathtt{R}$ there exists a position such that variables from $\mathbf{x}$ only appear at that position.
If we have a decomposer, $\mathcal{M}$ can be simplified to $\mathcal{M}[X/\mathbf{x}]$ that is obtained by substituting all variables in $\mathbf{x}$ by the same constant $X$ in $D_x$, $x\in\mathbf{x}$, then converting the result to normal form.
The partition function is calculated as $Z(\mathcal{M}) = [\left(Z(\mathcal{M}[X/\mathbf{x}])\right)^{|D_x|}$.

The next structural simplification comes from \emph{isolated variables}---one such variable in a predicate $\mathtt{R}$ at position $m$ is exclusive to $\mathtt{R}$ in all formulas containing $\mathtt{R}$.
Let $\mathbf{x}$ denote all isolated variables of $\mathtt{R}$ and $\mathbf{y}$ the rest of the variables, and $\mathbf{Y}_i\in D_\mathbf{y}$.
We obtain a simplified MLN $\mathcal{M}[\mathtt{R},\mathbf{x}]$ by generating the groundings of $\mathtt{R}(x, \mathbf{Y}_i)$ for $i=1,\ldots,|D_\mathbf{y}|$, deleting the formulas that evaluate $\mathtt{True}$ or $\mathtt{False}$, deleting all groundings of $\mathtt{R}$, and normalizing the result.
We get a combinatorial multiplier term to adjust the value of the partition function.
\begin{algorithm}[H]
\begin{algorithmic}
\REQUIRE A normal MLN $\mathcal{M}$
\ENSURE The value of the partition function $Z(\mathcal{M})$
\IF{$\mathcal{M}$ is fully ground out}
  \STATE Run Gibbs sampler to obtain $Z(\mathcal{M})$
  \RETURN $Z(\mathcal{M})$
\ENDIF
\IF{there exists a decomposer $\mathbf{x}$}
  \STATE Let $x\in\mathbf{x}$ and $X\in D_x$
  \RETURN LS$(\mathcal{M}[X/\mathbf{x}])^{|D_x|}$.
\ENDIF
\IF{there exists an isolated variable $x$}
  \RETURN $w(\mathtt{R})2^{p(\mathtt{R})}\prod_{i=1}^{|D_y|}{|D_x| \choose j_i}$LS($\mathcal{M}[\mathtt{R},x]$)
\ENDIF
\IF{exists singleton atom $\mathtt{R}$ that does not appear more than once in the same formula}
    \RETURN $\sum_{i=0}^{|D_x|}$ ${|D_x| \choose i}w(i)2^{p(i)}$LS($\mathcal{M}|\bar{\mathtt{R}}^i$).
\ENDIF
\STATE Choose an atom $A$
\RETURN $\sum_{\bar{A} \textrm{ in groundings of } A}$ LS($\mathcal{M}|\bar{A}$).
\end{algorithmic}
\caption{Lifted Sampling (LS) of an MLN}
\label{alg:liftedmln}
\end{algorithm}

The final simplification is known as the \emph{generalized binomial rule}, which relies on singleton atoms that do not appear more than once in the same formula.
Given such an atom $\mathtt{R}(x)$, we can simplify the MLN as $\mathcal{M}|\bar{\mathtt{R}}^i$, where $\bar{\mathtt{R}}^i$ is a truth assignment to all groundings of $\mathtt{R}$ such that exactly $i$ groundings are set to $\mathtt{True}$.
The simplified network is obtained by grounding all $\mathtt{R}(x)$ and setting all its groundings to match the assignment given by $\bar{\mathtt{R}}^i$, deleting the formulas that evaluate $\mathtt{True}$ or $\mathtt{False}$, deleting all groundings of $\mathtt{R}$, and normalizing the result.
We can compute the partition function by $Z(\mathcal{M}=\sum_{i=0}^{|D_x|}{|D_x| \choose i}Z(\mathcal{M}|\bar{\mathtt{R}}^i)w(i)2^{p(i)}$, where $w(i)$ is the exponentiated sum of the formulas that evaluate to $\mathtt{True}$, and $p(i)$ is the number of ground atoms that are removed when removing the formulas.

If we cannot find any heuristics, we have to resort to fully grounding out an atom, normalizing the result, and continuing with the remaining expressions.

\subsection{Probabilistic Inference Given Evidence}
If we look at probabilistic inference given evidence, at the level of the quantum protocol, this can be done in at least two ways:
First, to the Hamiltonian $H$ one can add some strong local ``clamping'' terms, effectively forcing some of the assignments to the desired values.
This is convenient from an implementation point of view, as it only requires few local changes in the Hamiltonian simulation procedure~\cite{berry2015simmulation} underlying the algorithm of~\cite{chowdhury2016quantum}.
However, it can be difficult to quantify the additional error due to the finite clamping strength and adding very strong clamping terms unfavorably affect the runtime of the algorithm.
Second, one can construct the local terms $h_l$ not from the full truth tables of the $f_j$, but instead use reduced truth tables given the evidence, to construct local terms $h_l$ that act non-trivially only on the grounded atoms for which no evidence exists.
This can only decrease the maximal weight (i.e., increase the temperature $1/\beta$), decrease the number of terms (in case some of them become completely trivial), and reduce the number of sites $n$.
Gibbs sampling with the new Hamiltonian is hence always at most as computationally costly as with the original Hamiltonian.

We can also use classical heuristics before employing the quantum protocol, as in the algorithm described in Section~\ref{sec:partfunc}. For first-order lifting methods, the presence of evidence is a problem, as it skews symmetries and potentially leads to a complete grounding out. To avoid this, \cite{venugopal2014evidence} proposed a distance function on the partially clamped network, and suggested a clustering to find clusters of similar groundings. All groundings in a cluster are replaced by their cluster center, reducing the overall network size to to $\mathcal{O}(r^c)$, where $r$ is maximum cluster size, compared to the original $\mathcal{O}(D^c)$. This in turn reduces $n$ in the overall complexity of the quantum Gibbs sampling protocol, as stated in Proposition~\ref{prop:complexity}.

\section{Conclusions and Future Work}
We hope that by fostering knowledge exchange between communities, for example concerning the typical properties of Gibbs distributions relevant for machine learning, progress towards more realistic and useful quantum algorithms can be made. In summary, we addressed the following aspects of probabilistic inference in MLNs:
\begin{itemize}
  \item We analyzed the computational complexity of the state-of-the-art quantum Gibbs sampling protocol given the structural properties of MLNs and we argued the theoretical limits of the approach. A term in the computational complexity reduces exponentially, albeit the overall complexity remains exponential in the number of nodes.
  \item Understanding the impact of the properties of the graph generated by an MLN on the computational complexity of quantum Gibbs sampling, we adapted a classical first-order lifting algorithm to reduce the complexity of the network. The algorithm mirrors lifted importance sampling, but instead of using a proposal distribution, it uses either classical MCMC or quantum Gibbs sampling.
  \item We studied the effects of evidence on quantum Gibbs sampling.
\end{itemize}

The protocols we considered rely on a universal quantum computer, which, given the hurdles in implementation, is still mainly of academic interest. We can, however, turn to methods that use current or near future quantum annealing devices, for instance, technology using quantum annealing with manufactured spins~\cite{johnson2011quantum,boixo2014evidence}. In this technology, the distribution of excited states after annealing follows approximately a Boltzmann distribution~\cite{adachi2015application}, albeit one has to pay attention to estimating persistent biases and the effective temperature estimation~\cite{perdomo-ortiz2015determination,benedetti2015estimation}.
This technology was used, for instance, for learning the structure of a Bayesian network~\cite{ogorman2014bayesian}, but the restricted connectivity between the spins causes difficulties for arbitrary graph structures, in contrast to the methods discussed here. Recent progress allows embedding arbitrary graphs, albeit at a quadratic cost in the number of spins in the worst-case scenario~\cite{zaribafiyan2016systematic,benedetti2016quantum}, and there is also a proposal for a quantum annealing architecture with all-to-all connectivity~\cite{lechner2015quantum}. Given the techniques described in this paper, it would be interesting to see whether we can achieve a scalable implementation with contemporary quantum annealing technologies, since machine learning demonstrations with this paradigm mainly focused on Boltzmann machines so far: MLNs have different topological features than Boltzmann machines, but they also have regularities that might allow an efficient embedding and subsequent inference.

\end{document}